\pdfoutput=1
\documentclass[11pt]{article}

\usepackage[final]{coling}
\usepackage{amssymb}
\usepackage{times}
\usepackage{latexsym}
\usepackage{amsmath}
\usepackage[T1]{fontenc}
\usepackage[utf8]{inputenc}

\usepackage{microtype}

\usepackage{inconsolata}

\usepackage{graphicx}
\usepackage{multirow}
\usepackage{paralist} 
\usepackage{subfigure} 
\usepackage{caption} 

\title{MPPO: Multi Pair-wise Preference Optimization for LLMs with Arbitrary Negative Samples}

\author{
    \textbf{Shuo Xie}, \textbf{Fangzhi Zhu}\thanks{Corresponding author.}, \textbf{Jiahui Wang}, 
    \textbf{Lulu Wen}, \textbf{Wei Dai}, \\
    \textbf{Xiaowei Chen}, \textbf{Junxiong Zhu}, \textbf{Kai Zhou}, 
    \textbf{Bo Zheng} \\
    Taobao \& Tmall Group of Alibaba \\
    \texttt{shuoxie090@gmail.com}; \\
    \texttt{\{fangzhi.zfz,wjh439451,wenlulu.wll,yiwei.dw\}@taobao.com} \\
    \texttt{\{qisheng.cxw,xike.zjx,kevin.zk\}@taobao.com};
    \texttt{bozheng@alibaba-inc.com}
}

\begin{document}
\maketitle
\begin{abstract}
Aligning Large Language Models (LLMs) with human feedback is crucial for their development. Existing preference optimization methods such as DPO and KTO, while improved based on Reinforcement Learning from Human Feedback (RLHF), are inherently derived from PPO, requiring a reference model that adds GPU memory resources and relies heavily on abundant preference data. 
Meanwhile, current preference optimization research mainly targets single-question scenarios with two replies, neglecting optimization with multiple replies, which leads to a waste of data in the application.
This study introduces the MPPO algorithm, which leverages the average likelihood of model responses to fit the reward function and maximizes the utilization of preference data. Through a comparison of Point-wise, Pair-wise, and List-wise implementations, we found that the Pair-wise approach achieves the best performance, significantly enhancing the quality of model responses.
Experimental results demonstrate MPPO's outstanding performance across various benchmarks. On MT-Bench, MPPO outperforms DPO, ORPO, and SimPO. Notably, on Arena-Hard, MPPO surpasses DPO and ORPO by substantial margins. These achievements underscore the remarkable advantages of MPPO in preference optimization tasks.

\end{abstract}

\section{Introduction}

As large language models (LLMs) advance at an impressive pace, their performance on various tasks approaches and even exceeds that of humans. 
Generally, the complete training process for a LLM entails three main stages: pre-training \cite{Brown2020LanguageMA}, task-specific fine-tuning (Supervised Fine-Tuning, SFT) \cite{Wei2021FinetunedLM, Wang2022SelfInstructAL} and preference optimization.
\begin{figure}
\centering
\includegraphics[width=3in]{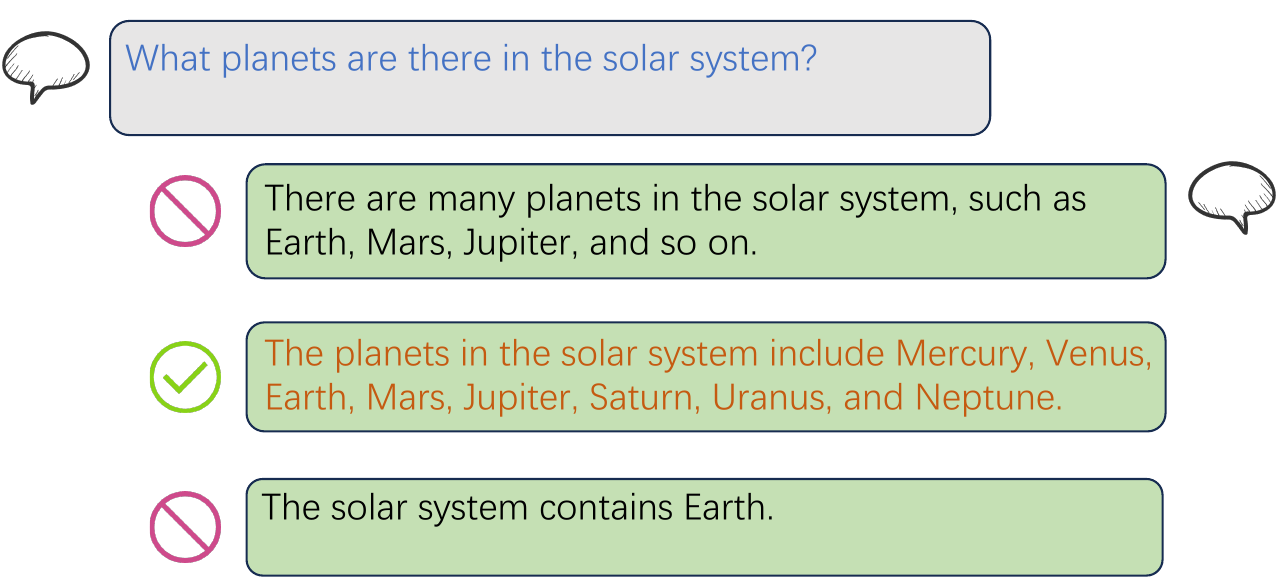}
% where an .eps filename suffix will be assumed under latex, 
% and a .pdf suffix will be assumed for pdflatex; or what has been declared
% via \DeclareGraphicsExtensions.
\caption{A simple example: When answering questions, the LLMs may generate multiple responses, but the quality of different responses varies.}
\label{fig:11}
\end{figure}

The pre-training phase involves unsupervised learning on large text datasets, which provides LLMs with a broad foundation of language knowledge \cite{Hu2022ASO, Almazrouei2023TheFS}. However, since pre-trained models typically learn general language patterns, their performance on specific tasks may be insufficient. Therefore, pre-trained LLMs often require further SFT to excel in practical applications.

The SFT process typically involves supervised learning, where the model is trained on a labeled dataset tailored to the target task \cite{gudibande2024the}. SFT enhances LLMs' performance on task-specific metrics, such as accuracy and relevance.
However, as depicted in Figure 1, SFT models may produce responses that diverge from human preferences when responding to queries \cite{Carlini2020ExtractingTD, pryzant-etal-2023-automatic}. Thus, an efficient preference optimization strategy is crucial for aligning their responses with human values and preferences. 

% Integrating human feedback has become a crucial method for calibrating LLMs. Reinforcement Learning from Human Feedback (RLHF) \cite{Christiano2017DeepRL,https://arxiv.org/abs/1706.03741} is a popular approach that adjusts language models for effective alignment and has provided a solid paradigm \cite{Ouyang2022TrainingLM} as exemplified by models like GPT4 \cite{Achiam2023GPT4TR}. Yet, the execution of this strategy poses challenges because it involves multiple training processes, and complete training necessitates the loading of several models, consequently consuming substantial GPU memory resources.

\begin{figure*}[t]
\centering
\includegraphics[width=6in]{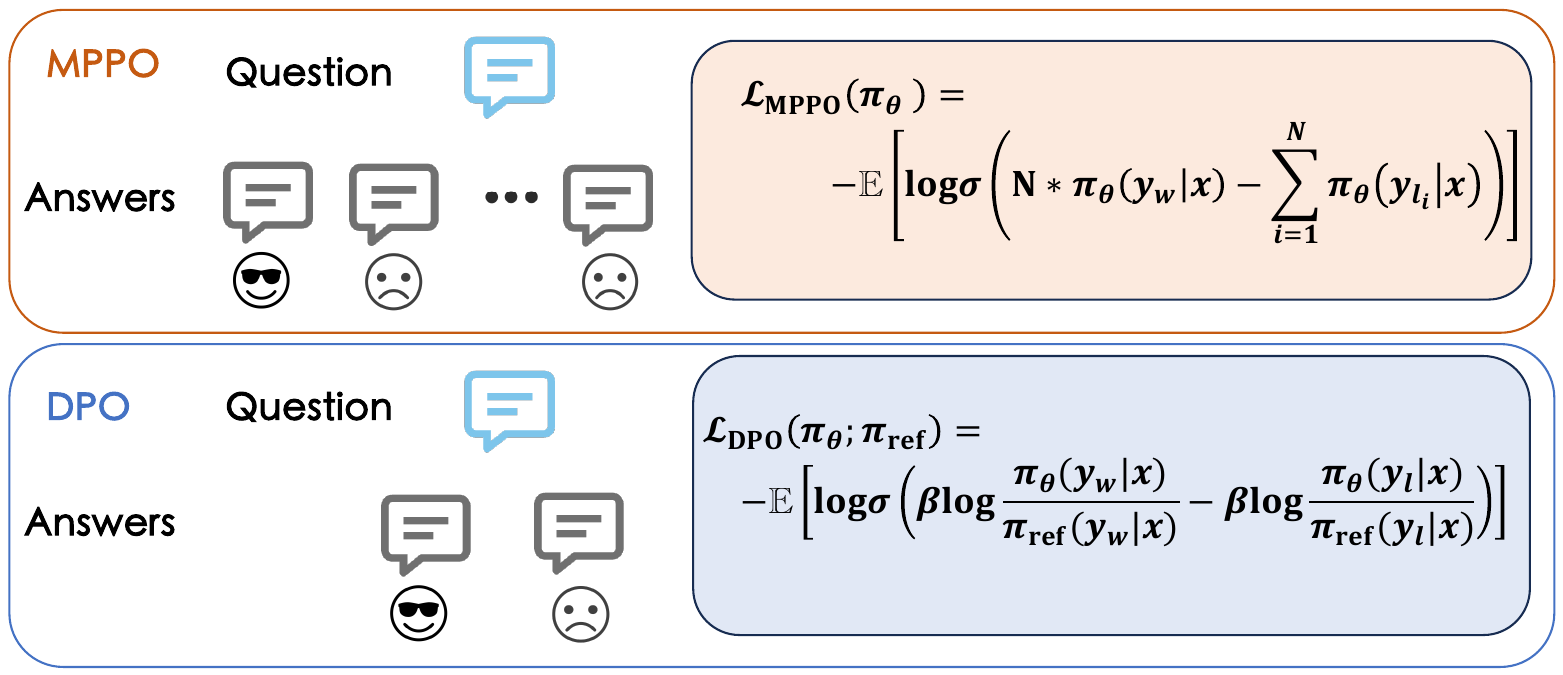}
% where an .eps filename suffix will be assumed under latex, 
% and a .pdf suffix will be assumed for pdflatex; or what has been declared
% via \DeclareGraphicsExtensions.
\caption{The primary differences between MPPO and DPO are three-fold: (1) MPPO directly models the reward function using the average likelihood of the responses. (2) MPPO can utilize any number of negative samples for training in sparse scenarios. (3) The optimization objective of MPPO does not require a reference model or any hyperparameters, making the model more stable.}
% \caption{The primary differences between MPPO and DPO are three-fold: (1) MPPO directly models the reward function using the average likelihood of the responses, so that after the model is fully trained, the output probability of responses with high rewards will be greater when sampling model responses. (2) MPPO can utilize any number of negative samples for training in sparse scenarios. (3) The optimization objective of MPPO does not require a reference model or any hyperparameters, making the model more stable.}
\label{fig:experiments}
\end{figure*}

% Recent studies on preference optimization, such as PPO and DPO \cite{Rafailov2023DirectPO} based on PPO and Bradley-Terry function, reparameterizing the reward function, integrating reward modeling with the preference learning phase, which reduces the training cost of the reward model during the preference alignment phase. However, there is a high data cost associated with densely fitting the reward function with preference data, making it less effective in data-sparse scenarios. Most preference optimization algorithms, such as DPOP \cite{Pal2024SmaugFF}, KTO \cite{Ethayarajh2024KTOMA}, and other optimizations based on DPO \cite{Rafailov2023DirectPO}, struggle to break free from the requirements of reference models. SimPO \cite{Meng2024SimPOSP} models the reward function through the generative process of preference pair, addressing the limitation of reference model, yet neglects the challenge of better modeling the reward function in data-sparse scenarios and fails to fully utilize all available preference information under such conditions. In real-world settings, obtaining partial human preference data can be relatively easy and efficient. For instance, extracting partial responses from the same prompt quickly is feasible \cite{Liu2024LiPOLP}, which prompts us to consider how to carry out preference optimization tasks within data-sparse environments.
Recent studies on preference optimization, including reinforcement learning with human feedback and direct preference optimization (DPO) \cite{Rafailov2023DirectPO}, have established effective methods for aligning language models. These approaches have proven successful, as exemplified by models like GPT-4 \cite{Achiam2023GPT4TR} \cite{Ouyang2022TrainingLM} and Llama-3 \cite{Dubey2024TheL3}.

Preference alignment methods have proven successful not only in aligning with human values but also across various downstream tasks such as enhancing factual accuracy \cite{Tian2023FinetuningLM, Cheng2024CanAA} and code-based question answering \cite{Gorbatovski2024ReinforcementLF}. However, as shown in Figure 2, existing preference alignment methods often require the use of reference models, and the optimization goals are not directly related to the generation of responses.
In this work, we innovatively introduce the use of the average likelihood value of model responses as an approximation for the reward function, named Multi Pair-wise Preference Optimization (MPPO). In the real world, we can generate multiple responses for the same query using different models, although these responses may not be dense. We refer to this as a sparse data scenario, which is more reflective of practical situations. We further explore how to better leverage the preference information from multiple preference samples. In summary, our work makes the following contributions:
\begin{itemize}
\item[$\bullet$] We propose a new algorithm that directly optimizes the policy model without training a reward model or relying on a reference model.
\item[$\bullet$] Our discussion on preference optimization for multiple responses to a single query in sparse data scenarios has led to conclusions that are more suitable for practical applications.
\item[$\bullet$] We thoroughly analyze three primary implementations of MPPO: Point-wise, Pair-wise, and List-wise. Among them, the Pair-wise approach achieves optimal performance, offering insights that can inspire the development of other preference optimization algorithms.
\item[$\bullet$] Our experiments on the UltraFeedback \cite{Cui2023UltraFeedbackBL} dataset demonstrate that the proposed algorithm outperforms previous methods on the MT-Bench and also surpasses algorithms like DPO and ORPO on Arena-Hard.
\end{itemize}

\section{Related Works}

In this section, we review the existing research on preference optimization for LLMs. Reinforcement learning from human feedback (RLHF) has been widely applied in the methods of preference optimization. In these methods, people first construct a reward model on preference dataset \cite{Casper2023OpenPA}, and then fine-tune the LLMs with reinforcement learning algorithms such as proximal policy optimization (PPO) \cite{Schulman2017ProximalPO} or its variants to maximize the estimated rewards. To enhance the stability of RLHF, Christiano et al. \cite{Christiano2017DeepRL} and Ouyang et al. \cite{Ouyang2022TrainingLM} proposed incorporating KL regularization based on the SFT model into preference optimization.

While PPO has achieved significant success in training high-performance prognostic models, this method requires prior training of a reward model \cite{Gao2022ScalingLF, Wang2024SecretsOR}. The reward model demands a large amount of dense data to ensure modeling accuracy, and the instability in training the reward model causes high sensitivity to hyperparameters \cite{Wang2024SecretsOR}. Completing the full training also necessitates loading multiple models, which consumes extensive GPU memory resources, thereby presenting a series of practical challenges.

To address the stringent need for a reward model, numerous scholars have pursued innovation \cite{Song2023PreferenceRO}. Recent studies, such as DPO \cite{Rafailov2023DirectPO}, have attempted to reparameterize the reward function by integrating reward modeling with the preference learning stage, thereby reducing the training costs of the preference optimization phase. Further research building on DPO, such as KTO \cite{Ethayarajh2024KTOMA}, circumvents the need for paired preference samples and allows for effective preference optimization even in the presence of imbalanced positive and negative samples. DPOP \cite{Pal2024SmaugFF} specifically optimizes against a potential issue encountered by DPO, that is, the diminishing likelihood of the model's assessment of preference examples during training, to prevent this occurrence. However, methods such as DPO, KTO \cite{Ethayarajh2024KTOMA}, and DPOP \cite{Pal2024SmaugFF} still rely on a KL regularizer centered around SFT, and complete training still demands the loading of both the policy model and reference model.

In response to this challenge, odds ratio preference optimization (ORPO) \cite{Hong2024ORPOMP} investigates the role and impact of SFT in model optimization with pairwise preference datasets, seamlessly integrating the SFT phase with the preference optimization stage and incorporating probability ratios into the optimization objective, thus eliminating the mandatory use of the reference model during the preference optimization phase. Simple preference optimization (SimPO) \cite{Meng2024SimPOSP} introduces length-normalized rewards and marginal target rewards, which similarly omit the reference model, rendering the preference optimization process both efficient and concise.

Our subsequent research will focus on how to effectively and concisely achieve preference optimization, as well as how to better model rewards in real-world scenarios with preference samples that include multiple responses.

\section{Methodology}
In this section, we first introduce the background of DPO. Then, we present the fundamental principle of our approach: fitting the reward function with the responses' average likelihood of the model. Extending from this principle, we have outlined three primary methods of implementation.
\subsection{Direct Preference Optimization}
DPO is a highly successful method for offline preference optimization. Compared to the RLHF training approach, DPO reparameterizes the reward function ${r(x,y)}$ using a closed-form expression with the optimal policy:
\begin{equation}
  % \label{eq:example}
r(x,y) = \beta \log\left(\frac{\pi_{\theta}(y|x)}{\pi_{\text{ref}}(y|x)}\right) + \beta \log Z(x),
\end{equation}
where $\pi_{\theta}$ is the policy model, $\pi_\text{ref}$ is the reference policy, typically the supervised SFT model, and $Z(x)$ is the partition function. Integrating the reward model into the Bradley-Terry objective:
\begin{equation}
  % \label{eq:example}
p(y_{w} > y_{l} | x) = \sigma(r(x,y_{w}) - r(x,y_{l})).
\end{equation}
DPO expresses the probability of preference data through a policy model rather than a reward model, thus generating the following objective:
\begin{multline}
\mathcal{L}_{\text{DPO}\left( \pi_{\theta}; \pi_{\text{ref}}\right)} = -\mathbb{E}_{(x,y_{w},y_{l})\sim \mathcal{D}} \\
\left[ \text{log} \sigma \left( \beta \text{log} \frac{\pi_{\theta}(y_{w}|x) }{\pi_{\text{ref}}(y_{w}|x) } -  \beta \text{log} \frac{\pi_{\theta}(y_{l}|x) }{\pi_{\text{ref}}(y_{l}|x) }\right)\right],
\end{multline}
where $(x,y_{w},y_{l})$ are preference pairs consisting of the query, the winning response, and the losing response from the preference dataset $\mathcal{D}$.
DPO integrates reward modeling with the preference learning stage, thus reducing the training costs of the preference optimization phase.

\subsection{MPPO: Fitting the Reward Function to the Average Likelihood of the Model Responses}
Using Eq. (3) as the optimization objective in DPO has three drawbacks: (1). It necessitates loading a reference model during training, which results in additional GPU memory requirements and computational costs. (2). Training data requires strictly positive and negative preference pairs. When dealing with multiple responses to a single query, this requirement leads to significant data redundancy and inefficiency. (3). DPO uses the policy model to represent the probability of preference data, but this does not fully correspond to the actual probability of the preference data.

To address the aforementioned issue, we proposed MPPO, which uses the average likelihood fitting of the model's responses as the reward function:
\begin{equation}
  % \label{eq:mppo_reward}
r_{\text{MPPO}}(x,y) =\prod_{i=1}^{|y|} {\pi_{\theta}(y_{i}|x,y_{<i})}^{\frac{1}{|y|}},\end{equation}
where $\pi_{\theta}(y_{i}|x,y_{<i})$ represents the probability of generating the $i$-th token $y_{i}$ given the input $x$ and the preceding tokens $y_{<i}$. Here, $|y|$ denotes the length of the response $y$, which is the total number of tokens in the response. This function calculates the geometric mean of the likelihoods for all tokens in the response. When the model is well-constructed, this approach increases the likelihood of generating better responses, guiding the model to favor producing superior responses.

\subsection{Three Primary Methods of Implementation}
The MPPO algorithm can be implemented in various ways, making it crucial to evaluate these methods to determine the most effective approach for aligning LLMs.

To more comprehensively study which aspects of preference data and implementation methods are most critical, we assume access to complete preference information and a diverse dataset that includes annotations for both high-quality and low-quality responses, as well as specific scores for each response (i.e., one query with $n$ responses and corresponding scores for those responses), we examine the effectiveness of different implementation strategies based on this premise.

There are three primary implementation approaches: Point-wise, Pair-wise, and List-wise. In the following discussion, we denote $p$ as:
\begin{equation}
  % \label{eq:example}
p =\prod_{i=1}^{|y|} {\pi_{\theta}(y_{i}|x,y_{<i})}^{\frac{1}{|y|}}.
\end{equation}
\subsubsection{Implementation Approach Based on Point-Wise}
The idea behind Point-wise approaches is that each query, response, and score is considered individually during training, rather than as pairs. This allows the score to be aligned with Point-wise predictions. Since the score takes on discrete values, this process can be treated as a multi-class classification problem. Optimization objectives can include cross-entropy loss and mean squared error, as described in Eq. (6) and Eq. (7), respectively.
% \begin{equation}
%   % \label{eq:example}
% L_{Cross-entropy} = -((score)*logp + (1-score)*log(1-p))
% \end{equation}
% \begin{multline}
% \mathcal{L}_{\text{Point-CE}}(\pi_{\theta}) = -\mathbb{E}_{(x,y)\sim \mathcal{D}}\\\left[ (score)*\text{log}p + (1-score)*\text{log}(1-p) \right]
% \end{multline}
% \begin{multline}
% \mathcal{L}_{\text{Point-MSE}}(\pi_{\theta}) = -\mathbb{E}_{(x,y)\sim \mathcal{D}}\\\left[ (score-p)^2 \right]
% \end{multline}
\begin{align}
\mathcal{L}_{\text{Point-CE}}(\pi_{\theta}) &= -\mathbb{E}_{(x,y)\sim \mathcal{D}} \Bigg[ \text{score} \cdot \log(p) +\nonumber \\
&\quad  (1 - \text{score}) \cdot \log(1 - p) \Bigg] ,\\
\mathcal{L}_{\text{Point-MSE}}(\pi_{\theta}) &= -\mathbb{E}_{(x,y)\sim \mathcal{D}} 
\left[ (\text{score} - p)^2 \right],
\end{align}
where score represents the reward value assigned to each response, and y can be both $y_{w}$ and $y_{l}$.
\subsubsection{Implementation Approach Based on Pair-Wise}
Pair-wise approaches focus on handling pairs of data or binary relations within datasets. At each instance, two response samples are selected—one designated as the positive sample and the other as the negative sample, based on their respective scores. The objective is to increase the average likelihood of selecting the positive sample while decreasing the likelihood of selecting the negative sample.

In the special case where each piece of data only contains one positive sample and one negative sample, the reward formula $r(x,y) = p$ can be substituted into the Bradley-Terry ranking objective $p(y_{w} > y_{l} | x) = \sigma(r(x,y_{w}) - r(x,y_{l}))$, resulting in $\text{Pair-Single}$ optimization function (8):
% \begin{multline}
% L_{\text{Pair-Single}}(\pi_{\theta}) = \\-\mathbb{E}_{(x,y_{w},y_{l})\sim D}\left[ \log \sigma (p_{w} - p_{l}) \right]
% \end{multline}
\begin{align}
\mathcal{L}_{\text{Pair-Single}}(\pi_{\theta}) &= -\mathbb{E}_{(x,y_{w},y_{l})\sim D} 
\left[ \log \sigma (p_{w} - p_{l}) \right].
\end{align}

However, in practical settings, obtaining large volumes of human preference data is comparatively easy and efficient. For example, a series of responses can be quickly generated from an identical query. Consequently, we consider the implementation of a Pair-wise approach when there are $N+1$ responses to the same prompt. A straightforward approach is to mark the response with the highest score value among the $N+1$ answers as the positive sample, and all the others as negative samples. This extends the Pair-Single method to two new variants: Pair-Multi-N-Separate (Pair-MNS) and Pair-Multi-N-Merge (Pair-MNM), as shown in Eq. (9) and Eq. (10).
\begin{multline}
\mathcal{L}_{\text{Pair-MNS}}(\pi_{\theta}) = -\mathbb{E}_{(x,y_{w},y_{l_{i}})\sim \mathcal{D}} \\ \left[ \sum_{i=1}^{N}{\underbrace{\log \sigma (p_{w} - p_{l_{i}})}_{\text{total of } {N \text{ items}}}} \right],
\end{multline}
\begin{multline}
\mathcal{L}_{\text{Pair-MNM}}(\pi_{\theta}) = -\mathbb{E}_{(x,y_{w},y_{l_{i}})\sim \mathcal{D}} \\
\left[ \log \sigma \left( N \cdot p_{w} - \sum_{i=1}^{N} p_{l_{i}} \right) \right].
\end{multline}

Subsequently, we adopted OpenAI's strategy for selecting data when training reward models, which involves randomly choosing a pair from $K$ data points for comparative training \cite{Ouyang2022TrainingLM}. In our method, we randomly select two data points from the $N+1$ responses and compare them based on their scores. The sample with the higher score is treated as the positive instance, while the one with the lower score is considered the negative instance. Based on this, we extend Equations (9) and (10) to incorporate this training framework as Pair-Multi-Combination-Separate (Pair-MCS) and Pair-Multi-Combination-Merge (Pair-MCM) in Eq. (11) and Eq (12):
% \begin{multline}
% \mathcal{L}_{{\text{Pair-MCS}}}(\pi_{\theta}) = -\mathbb{E}_{(x,y_{w},y_{l_{i}})\sim \mathcal{D}}\\ \left[ \sum_{i=1}^{N}{\underbrace{\log \sigma (p_{w} - p_{l_{i}})}_{\text{total of } {C_{N+1}^{2} \text{ items}}}} \right]
% \end{multline}
% \begin{multline}
% \mathcal{L}_{{\text{Pair-MCM}}}(\pi_{\theta}) = -\mathbb{E}_{(x,y_{w},y_{l_{i}})\sim \mathcal{D}} \\ \left[ \log \sigma \sum_{i=1}^{N}{\underbrace{(p_{w} - p_{l_{i}})}_{\text{total of } {C_{N+1}^{2} \text{ items}}}} \right]
% \end{multline}
\begin{multline}
\mathcal{L}_{\text{Pair-MCS}}(\pi_{\theta}) = -\mathbb{E}_{(x,y_{w},y_{l_{i}})\sim \mathcal{D}} \\
\left[ \sum_{i=1}^{N} \underbrace{\log \sigma (p_{w} - p_{l_{i}})}_{\text{total of } C_{N+1}^{2} \text{ items}} \right],
\end{multline}

\begin{multline}
\mathcal{L}_{\text{Pair-MCM}}(\pi_{\theta}) = -\mathbb{E}_{(x,y_{w},y_{l_{i}})\sim \mathcal{D}} \\
\left[ \log \sigma \left( \sum_{i=1}^{N} \underbrace{(p_{w} - p_{l_{i}})}_{\text{total of } C_{N+1}^{2} \text{ items}} \right) \right].
\end{multline}

\subsubsection{Implementation Approach Based on List-Wise}
Unlike Pair-wise methods, which focus on pairwise preferences between items, the List-wise approach considers the ordering of items across the entire list. This method aims to directly optimize ranking models based on a list's overall ranking quality. In this context, the list of responses contains $N+1$ items. Notably, when $N=1$, the List-wise and Pair-wise approaches are equivalent. A specific implementation of the List-wise method is List-MLE \cite{10.5555/3020751.3020798}. List-MLE applies Maximum Likelihood Estimation (MLE) directly to ranking list data, integrating Eq. (4) into the List-wise ranking loss to derive the List-MLE optimization function:
% \begin{multline} \mathcal{L}{\text{List-MLE}}(\pi{\theta}) = -\mathbb{E}{(x,y{1},\ldots,y_{N+1})\sim \mathcal{D}} \left[ \log \prod_{i=1}^{N+1}\frac{\exp(p_{i})}{\sum_{j=i}^{N+1}\exp(p_{j})} \right] \end{multline}
\begin{multline}
\mathcal{L}_{\text{List-MLE}}(\pi_{\theta}) = 
-\mathbb{E}_{(x,y_{1},\ldots,y_{N+1})\sim \mathcal{D}} \\\left[ \log\prod_{i=1}^{N+1}\frac{\exp(p_{i})}{\sum_{j=i}^{N+1}{\exp(p_{j})}}\right],
\end{multline}
where $p_{i}$ represents the reward value for the $i$-th item, and the items are arranged in descending order of their reward values.

% ListNet aims to minimize the discrepancy between the probability distributions of the predicted and true ranking outcomes, ensuring that the predictions closely resemble the actual results. The specific implementation formula is:
% \begin{equation}
% P(i) = \frac{exp(p_{i})}{\sum_{i=1}^{N+1}(exp(p_{i}))}
% \end{equation}
% \begin{equation}
% S(i) = \frac{exp(score_{i})}{\sum_{i=1}^{N+1}(exp(score_{i}))}
% \end{equation}
% \begin{equation}
% L_{List-ListNet} = -\sum_{i=1}^{N+1}(P(i)log(S(i) )
% \end{equation}

\section{Experimental Settings}

\subsection{Training Configurations}
\subsubsection{Model}
For training, we utilized the Llama3-8B model \cite{Dubey2024TheL3}, following the setups of Zephyr \cite{Tunstall2023ZephyrDD} and SimPO. The training process began by fine-tuning a foundational model on the UltraChat-200k dataset \cite{Ding2023EnhancingCL} to obtain a supervised fine-tuned (SFT) model. This SFT model then served as the initial model for preference optimization on the UltraFeedback dataset \cite{Cui2023UltraFeedbackBL}.

To ensure transparency, the SFT model was trained on open-source data, and we used the publicly available Llama3-8B-SFT weights from SimPO as our baseline. Preference optimization was conducted using three different implementations of MPPO: Point-wise, Pair-wise, and List-wise. Within each implementation, we experimented with different variations to determine the most effective alignment method for LLMs.

Unlike SimPO, which required extensive hyperparameter tuning, our approach only involved adjusting the learning rate, significantly reducing training costs while maintaining consistency and reliability.

% For training, we leveraged the Llama3-8B \cite{Dubey2024TheL3} model following the setups of Zephyr \cite{Tunstall2023ZephyrDD} and SimPO. Initially, training a foundational model on the UltraChat-200k \cite{Ding2023EnhancingCL} dataset to obtain a SFT model. Subsequently, employing this SFT model as a initial model, we conducted preference optimization on the UltraFeedback dataset \cite{Cui2023UltraFeedbackBL}. This setup afforded a high level of transparency since the SFT model was trained on open-source data. To maintain consistency with SimPO and enhance reliability, we utilized the publicly available Llama3-8B-SFT model weights from SimPO as our baseline model, and experimented with preference optimization on the UltraFeedback dataset utilizing three implementations of MPPO: Point-wise, Pair-wise, and List-wise. Notably, we also experimented with different approaches within each implementation modality to identify the most effective method for aligning LLMs. In contrast to SimPO, which required the search of multiple hyperparameters, our training necessitated merely the adjustment of the learning rate, thereby significantly reducing the cost of training.

\subsubsection{Datasets}
% We conducted preference optimization training on the UltraFeedback dataset, consisting of 64k instructions. For each instruction, four models were randomly selected to generate responses. GPT-4 scored each response on a scale of 1 to 10 based on criteria such as instruction adherence, authenticity, honesty, and helpfulness, where a higher score indicates a better response.
% 
% In the Point-wise implementation, each instruction along with its four responses and corresponding reward values was split into four separate samples for training, resulting in a total of 256k data points (64k*4). The reward values were normalized to a range between 0.1 and 1 by dividing by 10.
% 
% In the Pair-wise implementation, the response with the highest score was labeled as the positive sample, and one of the remaining three responses was randomly selected as the negative sample.
% 
% In the List-wise implementation, all four responses were sorted by their scores and trained together as a list.
We conducted preference optimization training on the UltraFeedback dataset, which includes 64k instructions. For each instruction, four models generated responses, and GPT-4 rated each response from 1 to 10 based on instruction adherence, authenticity, honesty, and helpfulness, with higher scores indicating better responses.

In the Point-wise implementation, each instruction with its four responses and reward values was split into four separate samples, totaling 256k data points (64k*4). Reward values were normalized to a range of 0.1 to 1 by dividing by 10.

In the Pair-Single implementation, the highest-scoring response was labeled positive, and one of the remaining three responses was randomly chosen as negative.

In the List-wise implementation, all four responses were sorted by their scores and trained together as a list.

\subsection{Leaderboard Evaluation}
To evaluate our model, we use two of the most popular open-ended instruction-following benchmarks: MT-Bench and Arena-Hard (details in Table 1). MT-Bench encompasses 80 tasks across 8 categories, with each task consisting of two rounds of question-and-answer phases. The newly released Arena-Hard is an enhanced version of MT-Bench that includes 500 clearly defined technical problem-solving queries. We report scores in accordance with the evaluation protocols of each benchmark. For MT-Bench, we provide the average score using GPT-4-Turbo-0409 as the evaluating model, which implements stricter grading criteria. For Arena-Hard, we report the win rate in comparison to the baseline model (GPT-4-0314). 
\begin{table*}
  \centering
  \begin{tabular}{lccccc}
    \hline
    \multicolumn{1}{c}{\textbf{}} & 
    \multicolumn{1}{c}{\textbf{Exs.}} & 
    \multicolumn{1}{c}{\textbf{Baseline Model}} & 
    \multicolumn{1}{c}{\textbf{Judge Model}}& 
    \multicolumn{1}{c}{\textbf{Scoring Type}}& 
    \multicolumn{1}{c}{\textbf{Metric}} \\
    \hline
    MT-Bench & 80 & - & GPT-4-Turbo-0409 & Single-answer grading & Rating of 1-10 \\
    Arena-Hard & 500 & GPT-4-0314 & GPT-4-Turbo-0409 & Pairwise comparison & Win rate \\
    \hline
  \end{tabular}
  \caption{Comparison of baseline and judge models on MT-Bench and Arena-Hard datasets.}
  \label{tab:model-comparison}
\end{table*}

\section{Results and Analysis}
In this section, we present the results of our experiments in Section 5.1, which include the outcomes for various implementations of MPPO and comparisons with SOTA preference optimization algorithms such as DPO, KTO, ORPO, and SimPO. In the subsequent subsections, 5.2 and 5.3, we draw several conclusions regarding preference optimization by analyzing and comparing these experimental results.

We have primarily explored four Research Questions (RQs) regarding the MPPO method:
\begin{itemize}
  \item \textbf{RQ1:} Are all implementations of MPPO: Point-wise, Pair-wise, and List-wise effective? Which method is the most effective?
  \item \textbf{RQ2:}  Is the use of multiple samples ($N+1$) preferable to optimizing with just a single positive and a single negative sample?
  \item \textbf{RQ3:} In sparse data scenarios, is the optimization goal of collaboratively leveraging multiple samples for preference optimization effective?
  \item \textbf{RQ4:} In sparse data scenarios with multiple responses, is it necessary to consider multiple samples for reward fitting (MCM), or is it sufficient to focus on just one optimal response (MNM)?
\end{itemize}

\subsection{Main Results}
In Table 2, we present the results of various implementations of MPPO and several preference optimization algorithms on MT-bench and Arena-hard benchmarks. It is observable that while all algorithms yield certain performance gains in the preference optimization for the SFT model, the magnitude of these improvements varies. The Pair-MNM implementation of MPPO achieved the highest scores on the MT-Bench leaderboard, surpassing the SFT model and the DPO, SimPO by 1.54 points, 0.23 points, and 0.19 points, respectively, demonstrating a significant enhancement and establishing itself as the latest SOTA algorithm.

In the Arena-Hard evaluation, the Pair-MNM implementation of MPPO placed second with a win rate of 21.6, only trailing behind SimPO, which had a win rate of 23.4. Pair-MNM outperformed DPO (win rate of 15.9), KTO (12.8), and ORPO (10.7).

\begin{table}
  \centering
  \begin{tabular}{lccccc}
    \hline
    \multicolumn{1}{c}{\textbf{Method}} & 
    \multicolumn{1}{c}{\textbf{Mt-Bench}} & 
    \multicolumn{1}{c}{\textbf{Arena-Hard}} \\
    \hline
    % \multicolumn{1}{c}{\textbf{}} & 
    % \multicolumn{1}{c}{\textbf{Leadboard Score}} & 
    % \multicolumn{1}{c}{\textbf{Leadboard Score}} & \\
    % \hline
    {SFT} & 4.62  & 3.3  &\\
    \hline
    {DPO} & 5.93  & 15.9 &  \\
    {KTO} & 5.87  & 12.8 &  \\
    {ORPO} & 5.49  & 10.7 &  \\
    {SimPO} & 5.97  & \textbf{23.4} &  \\
    {Point-CE} &    4.38&  12.8&  \\
    {Point-MSE} &   4.43 & 13.1 &  \\
    {Pair-Single} & 5.96  & 19.1 &  \\
    {Pair-MNS} & 5.84  & 5.1 &  \\
    {Pair-MNM} & \textbf{6.16}  & 21.6 &  \\
    {Pair-MCS} & 5.72  & 14.6 &  \\
    {Pair-MCM} & 5.77  & 7.8 &  \\
    {List-MLE} & 5.87  &  5.4&  \\
    \hline
  \end{tabular}
  \caption{The results of MT-Bench and Arena-Hard. The SFT models are trained on the UltraChat dataset, and then preference optimization models are trained from the SFT models using algorithms such as DPO, MPPO, etc.}
  \label{tab:model-comparison}
\end{table}

\subsection{Comparison of Three Implementation Approaches: Point-wise, Pair-wise, and List-wise (\textbf {RQ1})}
Based on the analysis results in Table 2, while MPPO algorithm has various implementation strategies, not all strategies can achieve effective goals. Firstly, among the three MPPO implementation strategies, the Pair-wise method stands out, outperforming the List-wise and Point-wise methods in both MT-Bench and Arena-hard benchmarks.

The Point-wise method, despite high expectations, did not perform as well as anticipated. In fact, it underperformed compared to the original SFT model on the MT-Bench evaluation set. This suggests that relying solely on the magnitude of label information is inadequate and that incorporating preference information is crucial. Figure 3 illustrates the issue with the Point-wise method's Point-CE training: the scores for all answers remain relatively high (between 0.1 and 1) throughout the training period. Consequently, the average likelihood of each answer increases uniformly. However, the score difference between the best response and several poor responses decreases, making it harder to distinguish high-quality response from lower-quality ones. This explains why the Point-CE model performs worse than the SFT model.

Additionally, it is important to note that there is only a positive correlation, not an exact correspondence, between the responses' average likelihood and reward values. Therefore, directly approximating these specific values may not be effective. The inherent randomness of scores generated by GPT-4 also adds complexity to the modeling process.

% \begin{figure}[!t]
% \includegraphics[width=3in]{image33.pdf}
% % where an .eps filename suffix will be assumed under latex, 
% % and a .pdf suffix will be assumed for pdflatex; or what has been declared
% % via \DeclareGraphicsExtensions.
% \caption{Training Logs for Point-CE (N=3): The left figure shows the change in average likelihood for the chosen and three rejected samples; the right figure illustrates the overall loss variation during the training period.}
% \label{fig:2}
% \end{figure}

\begin{figure}[!t]
  \centering
    \subfigure[]
    {
    	\begin{minipage}{3.6cm}
    	\centering
            \label{fig:re0}
    	\includegraphics[width=\textwidth]{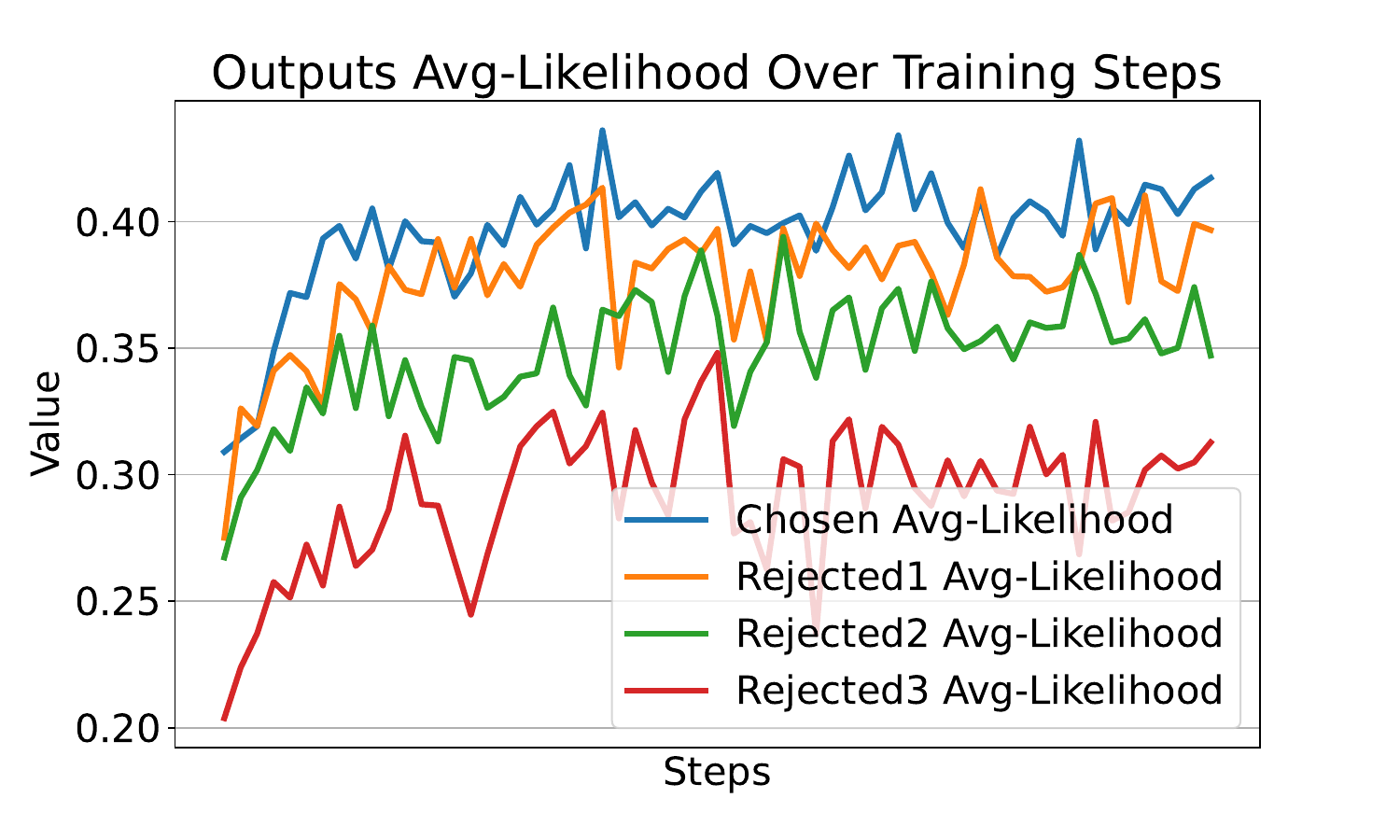}
    	\end{minipage}
     }\subfigure[]
    {
    	\begin{minipage}{3.6cm}
    	\centering
            \label{fig:loss0}
    	\includegraphics[width=\textwidth]{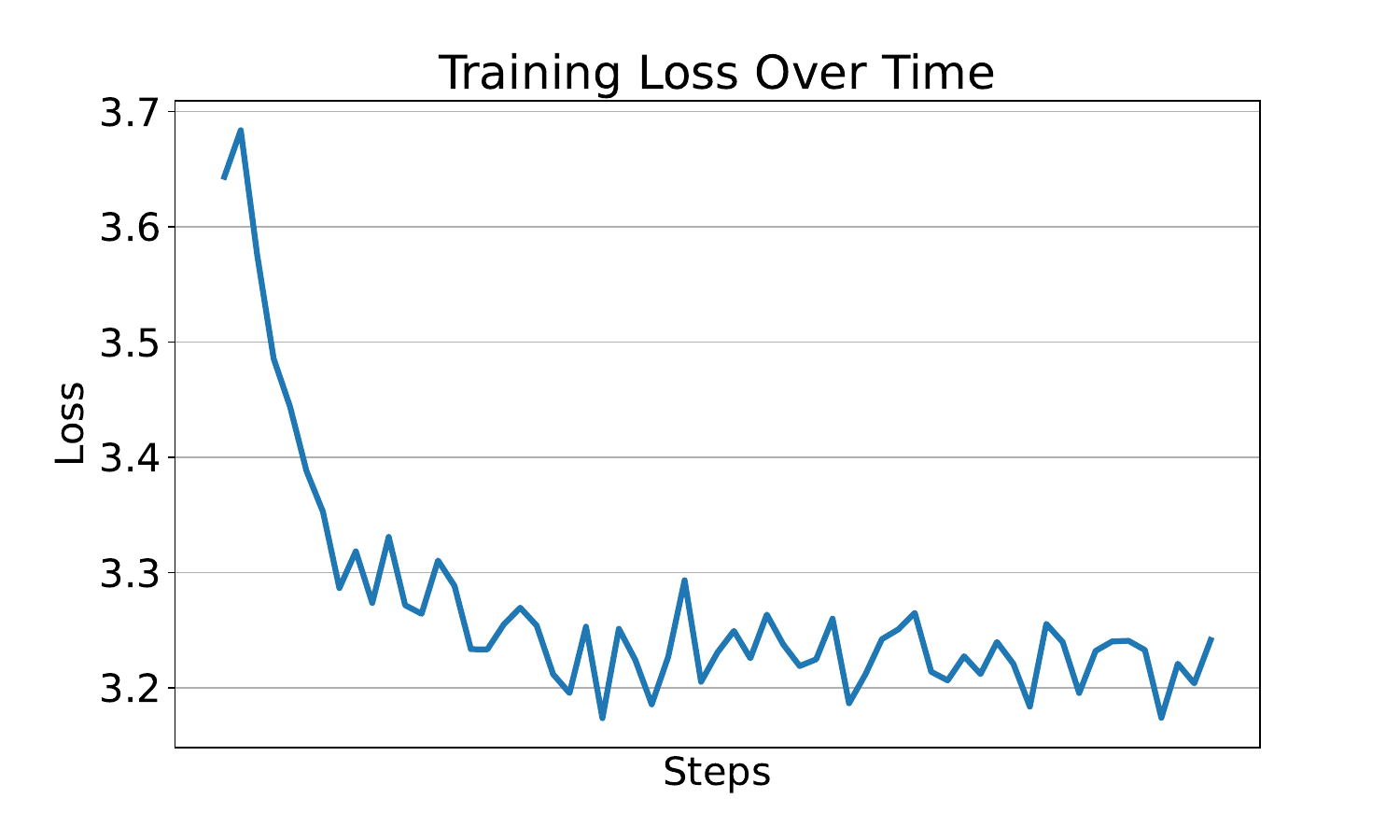}
    	\end{minipage}
     }
  \caption{Training Logs for Point-CE (N=3). (a) shows the variation in average likelihood for the chosen and three rejected samples; (b) illustrates the overall loss variation during the training period.}
  \label{fig:2}
\end{figure}

In conclusion, the List-wise approach exhibits certain disadvantages compared to Pair-wise methods. While it performs well on the MT-Bench benchmark, its performance is poor on Arena-Hard. This indicates the instability of the List-wise method, as well as the higher level of challenge presented by the Arena-Hard evaluation set compared to the MT-Bench. Although List-wise approach take preference information into account, it focus on preference ranking and do not strongly reflect preference information. Therefore, the list ranking information can be considered as weak preference information.

\subsection{Comparison of Implementation Methodology Based on Pair-wise}
In various implementations based on the Pair-wise strategy, the Pair-MNM approach achieved the best evaluation results. Subsequently, we will compare other Pair-Wise-based implementations to analyze the reasons behind the superior performance of Pair-MNM.
% \begin{figure}[!t]
% \includegraphics[width=3in]{image44.pdf}
% % where an .eps filename suffix will be assumed under latex, 
% % and a .pdf suffix will be assumed for pdflatex; or what has been declared
% % via \DeclareGraphicsExtensions.
% \caption{Based on the pair-wise implementation approach, the training records (N=3) are listed from top to bottom as follows:
% (a). Pair-Single
% (b). Pair-MNS
% (c). Pair-MNM
% (d). Pair-MCM}
% \label{fig:3}
% \end{figure}

\begin{figure}[!t]
  \centering
     \subfigure[]
    {
    	\begin{minipage}{3.6cm}
    	\centering
            \label{fig:re1}
    	\includegraphics[width=\textwidth]{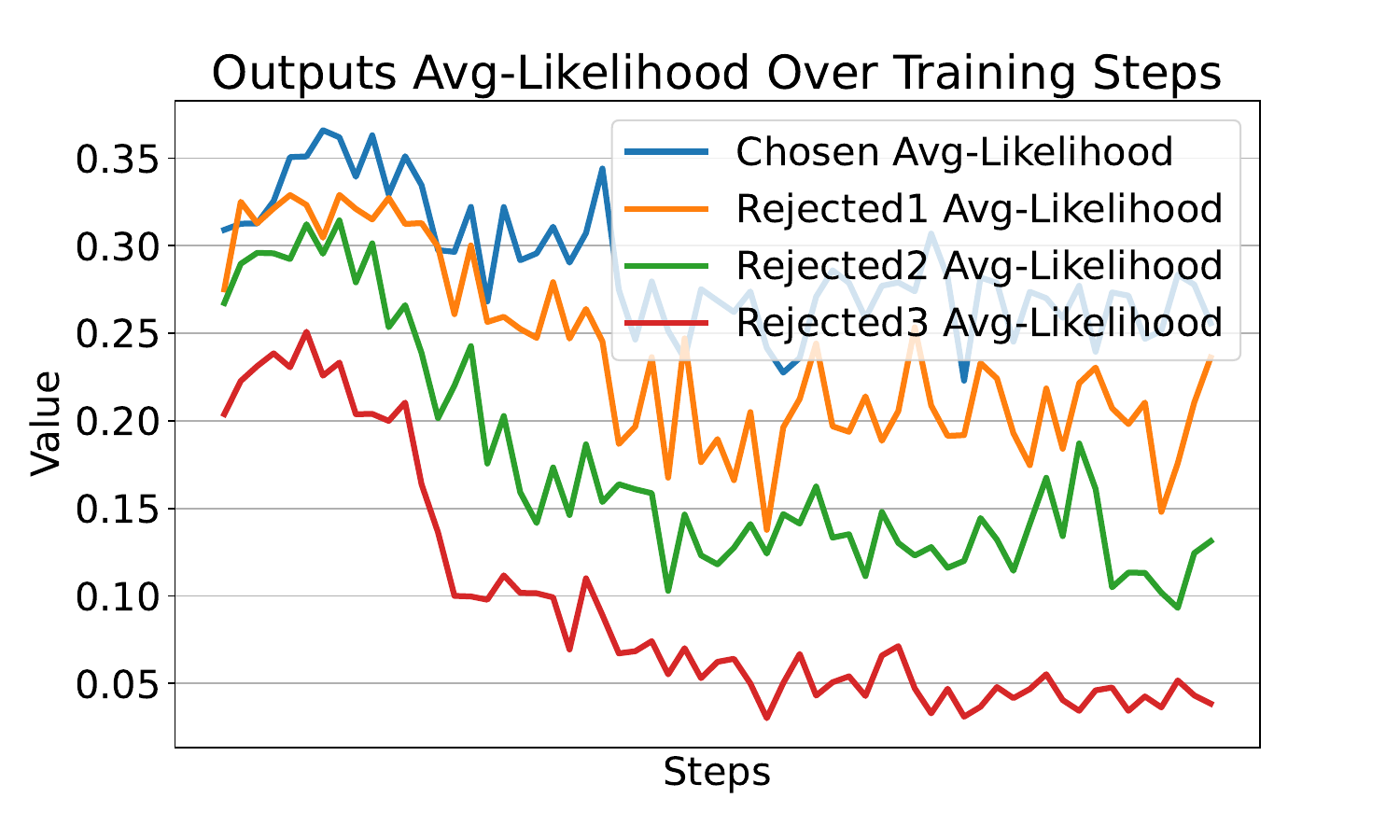}
    	\end{minipage}
     }\subfigure[]
    {
    	\begin{minipage}{3.6cm}
    	\centering
            \label{fig:loss1}
    	\includegraphics[width=\textwidth]{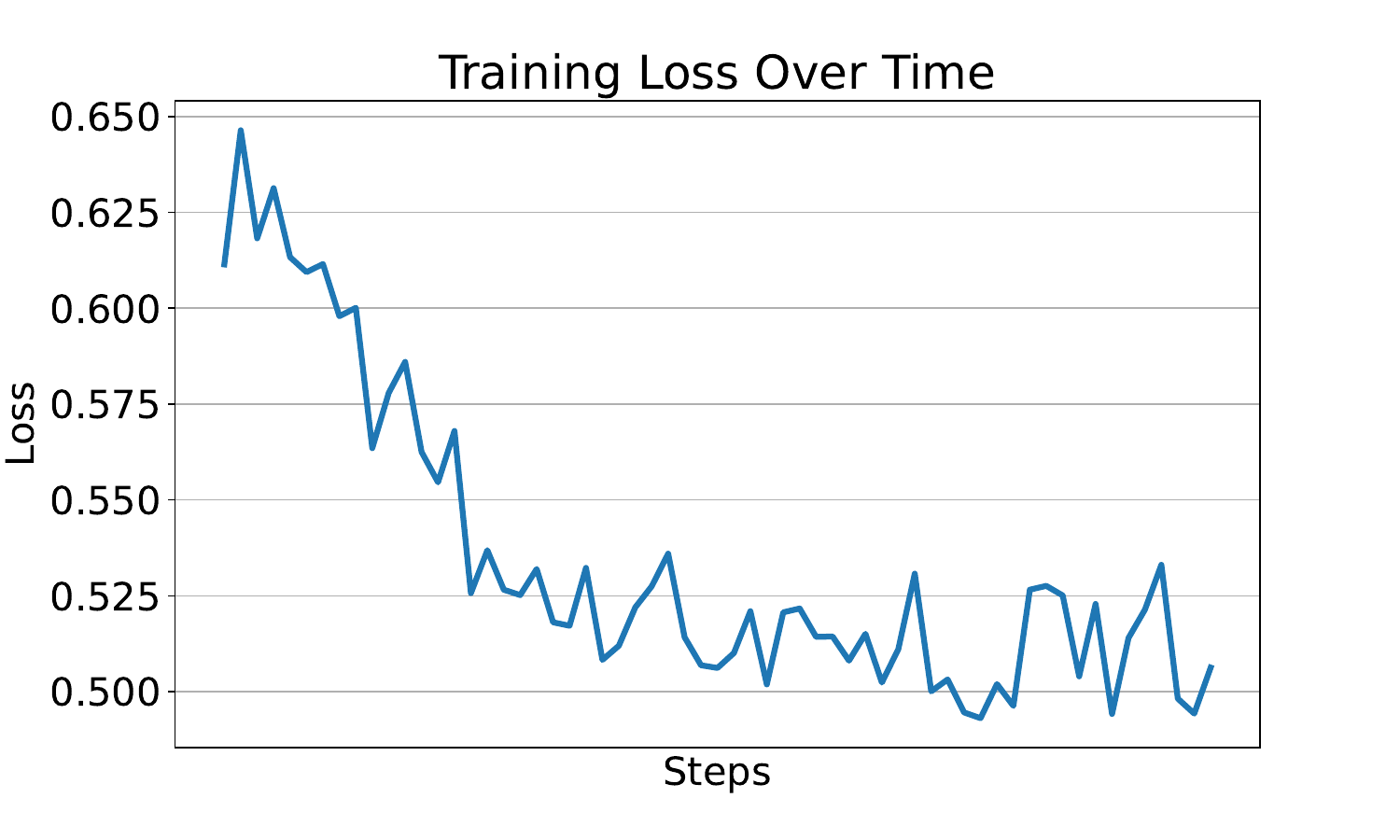}
    	\end{minipage}
     }
     \subfigure[]
    {
    	\begin{minipage}{3.6cm}
    	\centering
            \label{fig:re2}
    	\includegraphics[width=\textwidth]{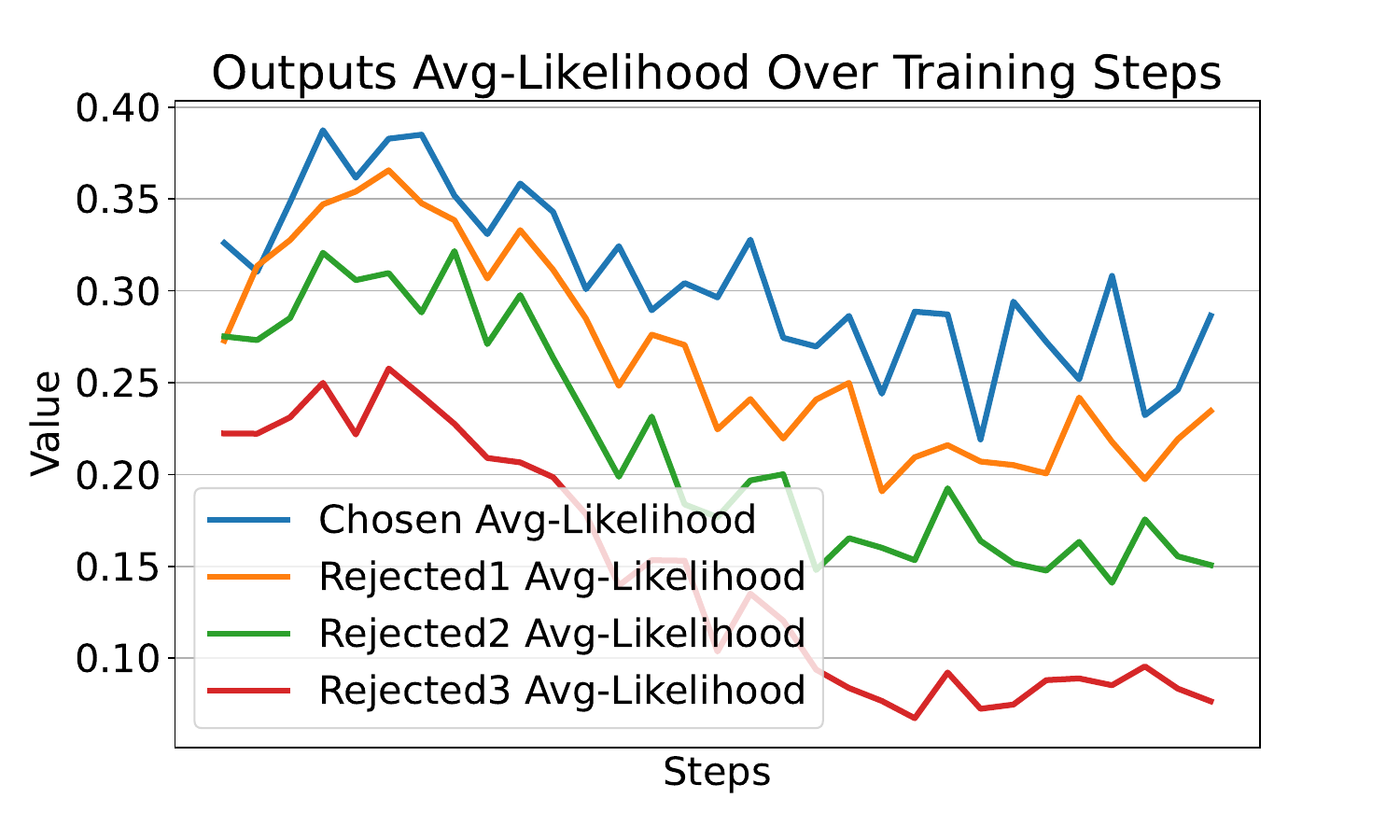}
    	\end{minipage}
     }\subfigure[]
    {
    	\begin{minipage}{3.6cm}
    	\centering
            \label{fig:loss2}
    	\includegraphics[width=\textwidth]{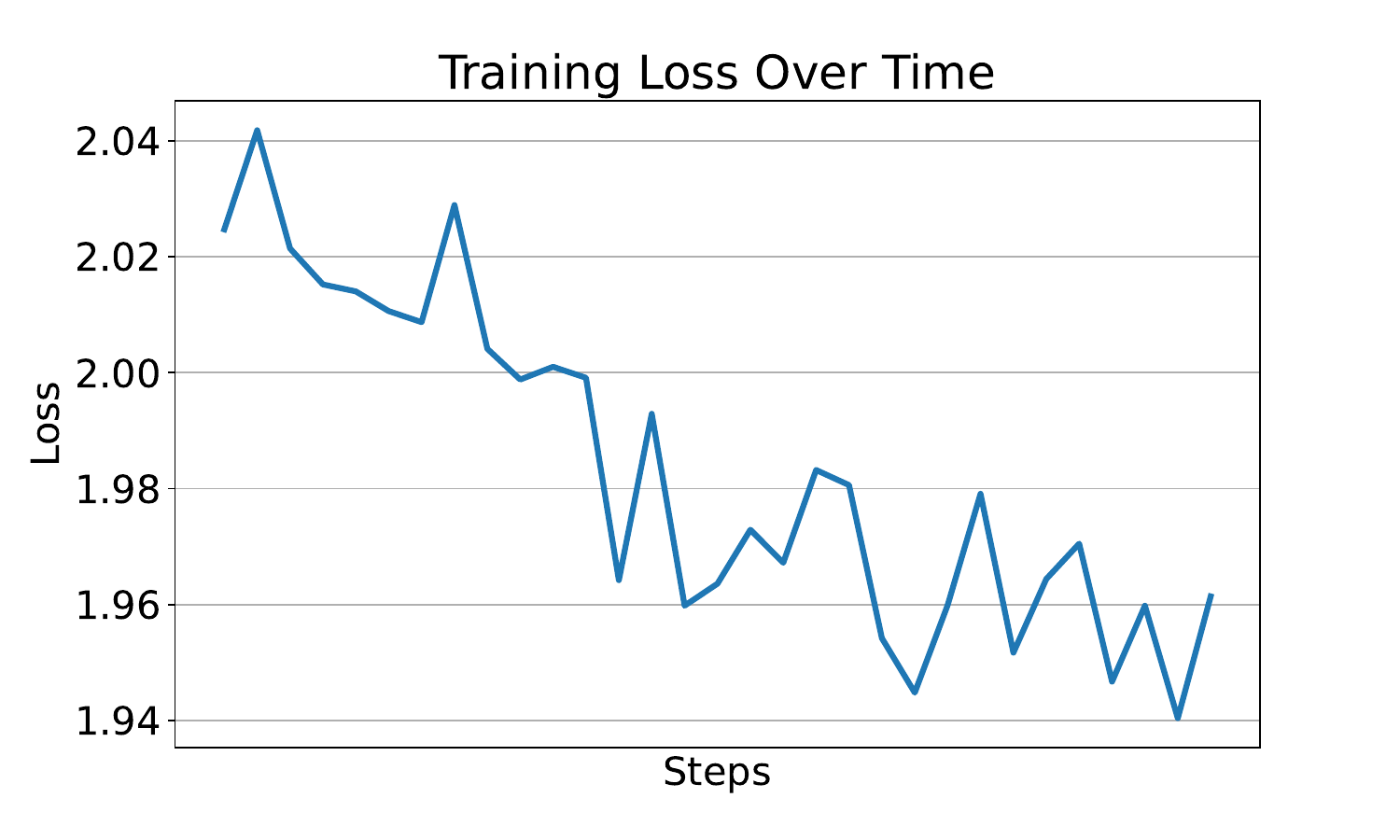}
    	\end{minipage}
     }
     \subfigure[]
    {
    	\begin{minipage}{3.6cm}
    	\centering
            \label{fig:re3}
    	\includegraphics[width=\textwidth]{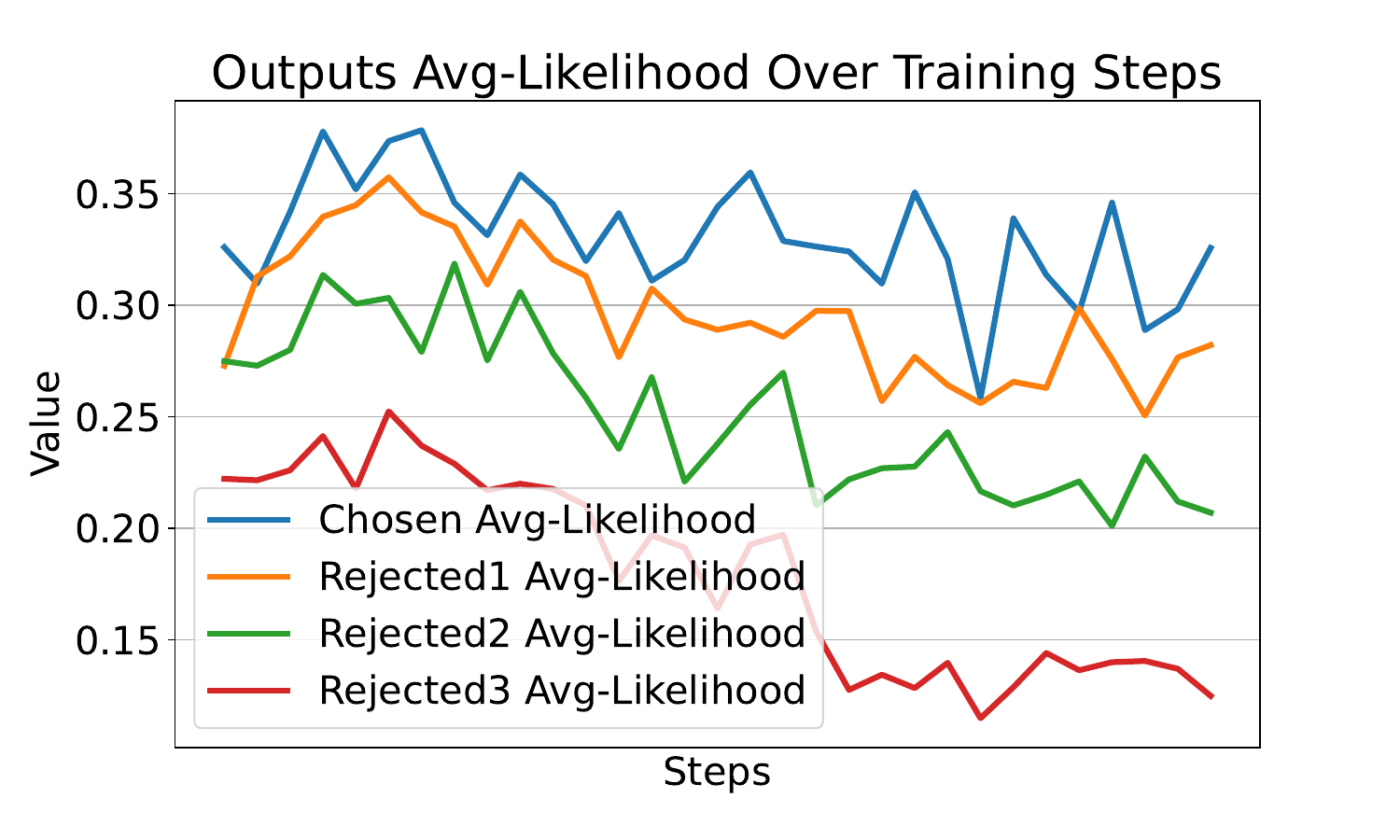}
    	\end{minipage}
     }\subfigure[]
    {
    	\begin{minipage}{3.6cm}
    	\centering
            \label{fig:loss3}
    	\includegraphics[width=\textwidth]{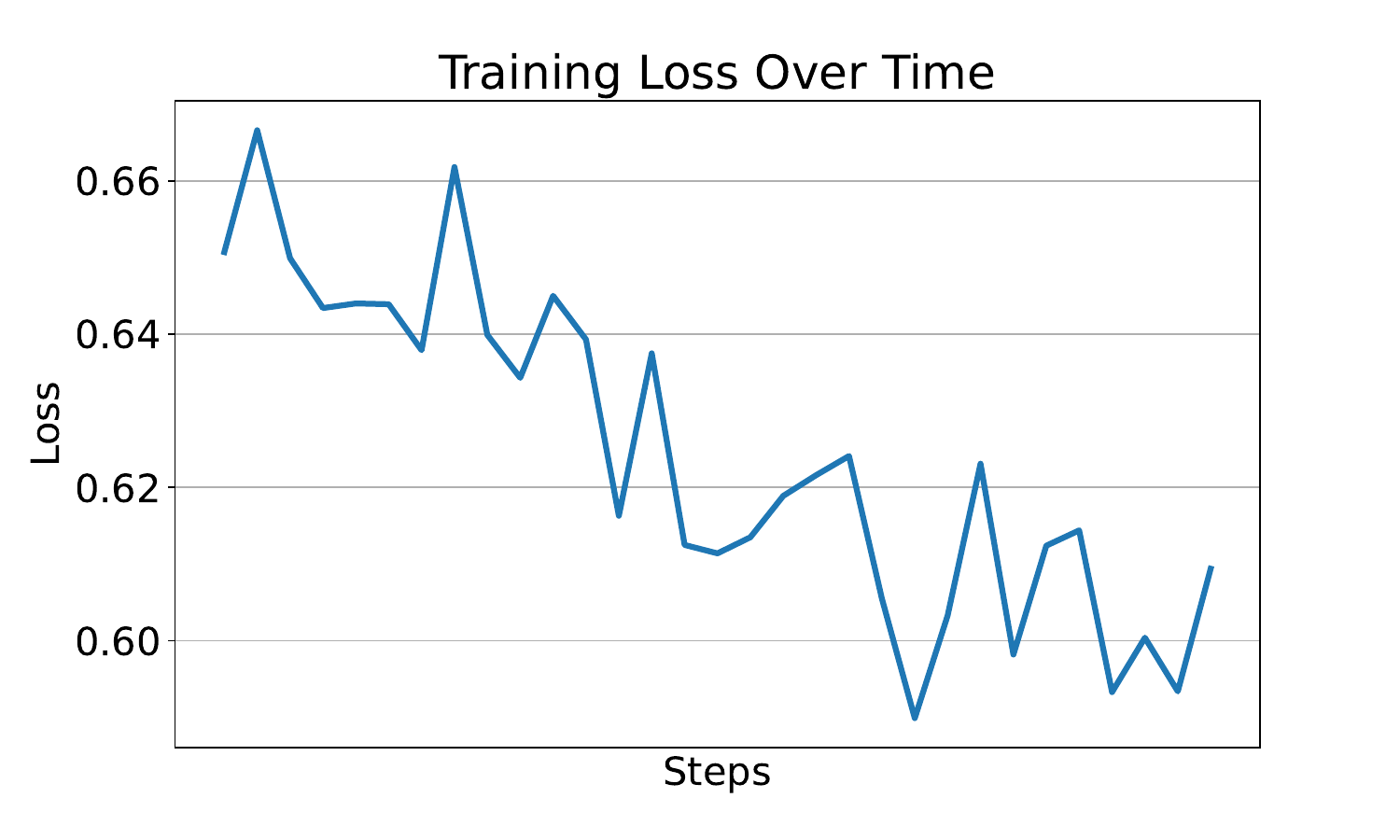}
    	\end{minipage}
     }
     \subfigure[]
    {
    	\begin{minipage}{3.6cm}
    	\centering
            \label{fig:re4}
    	\includegraphics[width=\textwidth]{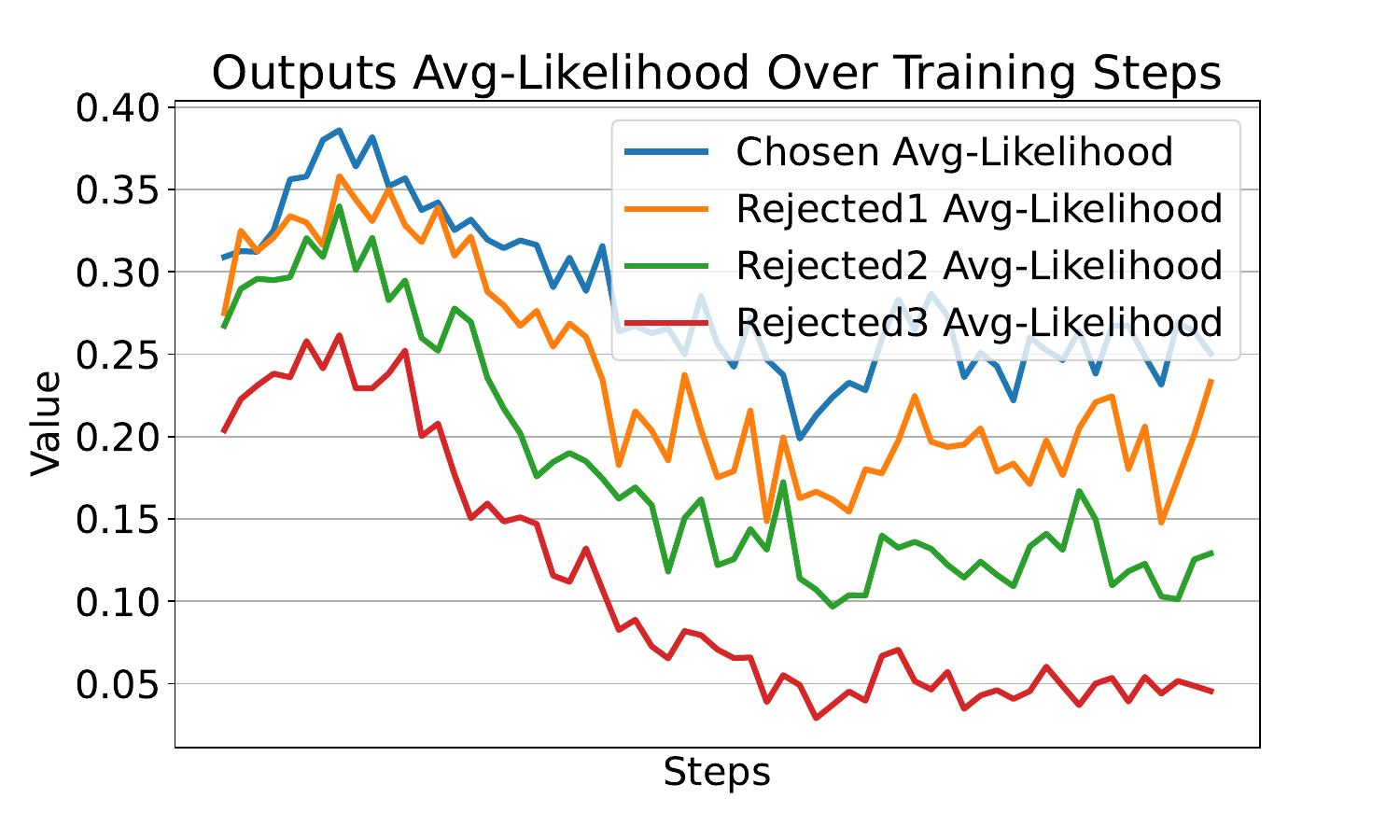}
    	\end{minipage}
     }\subfigure[]
    {
    	\begin{minipage}{3.6cm}
    	\centering
            \label{fig:loss4}
    	\includegraphics[width=\textwidth]{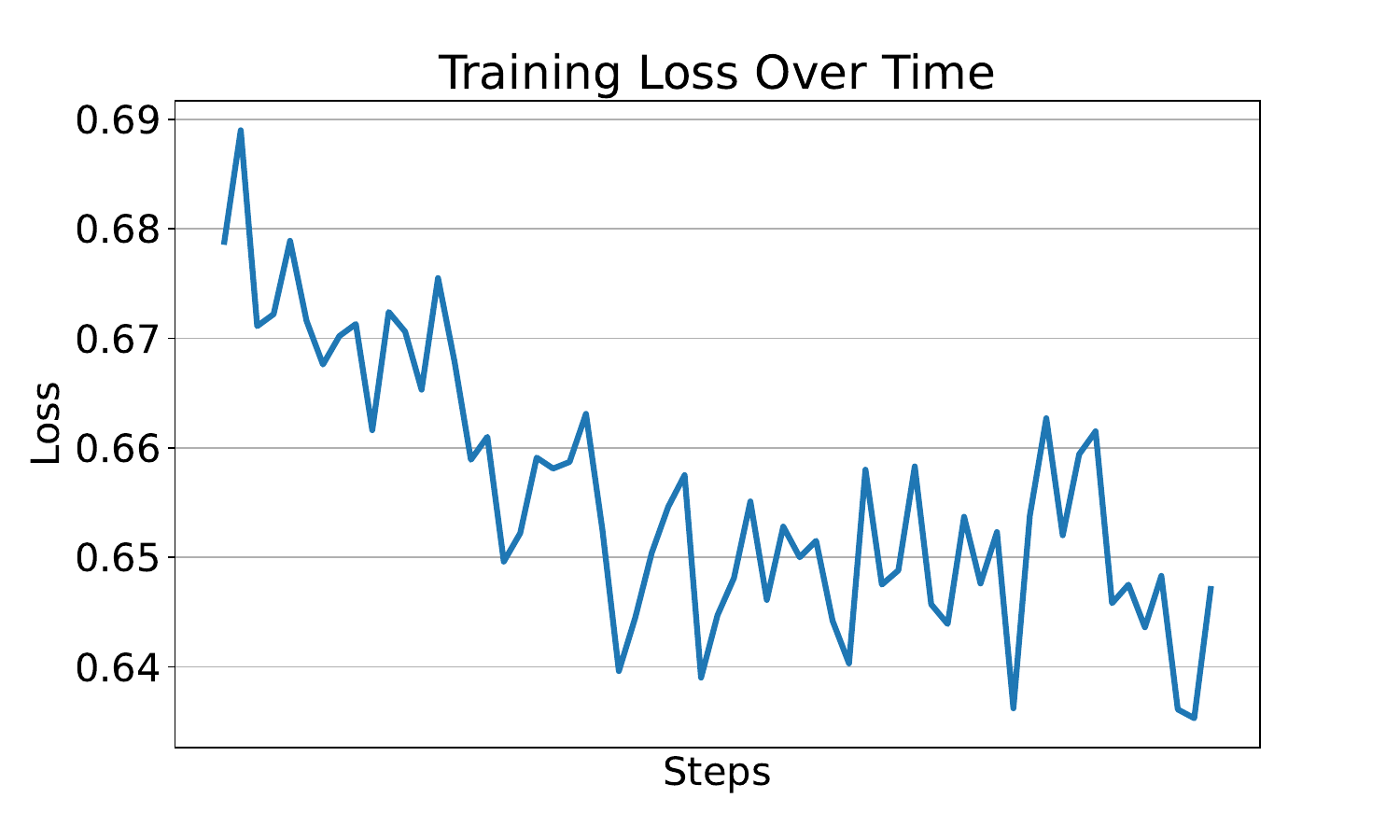}
    	\end{minipage}
     }
    \caption{Based on the Pair-wise implementation approach, the training records (N=3) are organized and listed from top to bottom as follows:
(a) and (b). Pair-Single implementation;
(c) and (d). Pair-MNS implementation;
(e) and (f). Pair-MNM implementation;
(g) and (h). Pair-MCM implementation.}
\label{fig:3}
\end{figure}

% \subsubsection{Multi-Pair-N-Merge vs Single-Pair: Considering all responses will enhance the preference model}
    \textbf{Considering all responses will enhance the preference model. (RQ2)} Compared to Single-pair optimization, the Pair-MNM approach evaluates all samples collectively. It designates the highest-quality answer as the positive sample and classifies the remaining responses as negative. This method enhances the incorporation of preference information by focusing on increasing the average likelihood of the chosen answers and decreasing that of the rejected ones. As shown in Figures 4a and 4c, Pair-MNM effectively amplifies the distinction between the chosen answer and each rejected one throughout the training process. This results in improved model performance, as it better learns to differentiate between varying responses.
    % 在此处添加更多条目

    \textbf{Incorporating Multiple Rejections Collaboratively for Optimal Objective Achievement in Preference Optimization. (RQ3)} Both Pair-MNM and Pair-MNS approaches, while based on the Single-pair method and considering multiple rejections, differ in how they handle these rejections. Pair-MNS calculates the average likelihood difference between the chosen instance and each rejected instance separately, summing the log-sigmoid values of these differences. This means each rejected instance independently influences the optimization objective. In contrast, Pair-MNM considers all rejected instances collectively, aiming to synergistically account for their combined impact to optimize the overall objective.

    The subtle differences in the optimization objectives lead to different outcomes during training, as depicted in Figures 4 c-f. While the loss curves and average likelihood trends are fundamentally similar, Pair-MNM consistently achieves an average likelihood of approximately 0.05 higher than Pair-MNS. This suggests that Pair-MNM, by integrating the discrepancies between multiple rejections relative to the chosen instance, results in a more effective decision boundary.

    \textbf{Just need one Optimal Response. (RQ4)} The main difference between Pair-MNM and Pair-MCM lies in how positive and negative samples are selected in sparse data scenarios. Pair-MNM chooses the highest-scoring sample as positive and labels all others as negative. In contrast, Pair-MCM uses a method similar to OpenAI’s approach: it draws two samples from N+1 data points, compares them based on their scores, and labels the higher-scoring sample as positive.

    Unlike Pair-MNM, which treats all rejected samples equally, Pair-MCM adjusts suppression intensity based on each sample's score, applying greater suppression to lower-scoring samples and less to higher ones. As illustrated in Figure 4g, this approach sometimes results in incorrect promotion of samples relative to the positive sample, as seen in the mean likelihood trends.

    Our experiments show that Pair-MCM does not perform better in sparse data scenarios. Thus, for multiple-answer preference optimization in such contexts, focusing on a single optimal response (Pair-MNM) and suppressing other samples is more effective for achieving better model performance and preference optimization.

\section{Conclusion}
% In this article, we introduce a preference optimization algorithm for direct modeling reward models in sparse data scenarios, namely: MPPO (Multi Pair-wise Preference Optimization). By comparison with various existing preference optimization methods, MPPO has achieved superior performance on the Llama3-8B model. Furthermore, we compared various implementations of MPPO, concluding that: Pair-wise implementations outperform Point-wise and List-wise implementations, considering all responses enhances the preference model, collaborative consideration of multiple rejections leads to optimal optimization results, and merely one optimal response is needed without the necessity of multiple sampling for preference pairs. MPPO provides an excellent example and insight into how to utilize preference data with multiple responses to single questions and addresses common issues encountered in real-world applications under sparse data conditions.

In this article, we present MPPO (Multi Pair-wise Preference Optimization), a preference optimization algorithm designed for directly modeling reward models in sparse data scenarios. Compared to existing methods, MPPO demonstrates superior performance on the Llama3-8B model. Our analysis reveals that Pair-wise implementations outperform Point-wise and List-wise approaches. Additionally, considering all responses enhances the preference model, and collaboratively addressing multiple rejections yields optimal results. Notably, only one optimal response is needed, eliminating the need for multiple sampling of preference pairs. MPPO effectively illustrates how to leverage preference data from multiple responses to a single query and addresses common challenges in real-world sparse data applications.
\section*{Limitations and Ethics}
Our work primarily proposes a preference optimization algorithm for directly modeling reward models in sparse data scenarios, without the need for a reference model. We acknowledge that the main limitations of this study are as follows:
% Our work proposes a preference optimization algorithm for directly modeling reward functions in sparse data scenarios, without needing a reference model. The main limitations of this study are:

1. We covered various implementation methods, including Point-wise, Pair-wise, and List-wise, as well as others like logistic ranking loss and ListNet. Future work will explore a broader range of methods and analyze their strengths and weaknesses in more detail.

2. We did not clearly define the boundary between data-rich and data-scarce scenarios in preference optimization. This will be addressed in future work through further discussion and experimental analysis.

All experiments are conducted on publicly available datasets; no scientific ethical violations or privacy infringements occurred.
\section*{Acknowledgments}

\bibliography{custom}

% \appendix

% \section{Example Appendix}
% \label{sec:appendix}

% This is an appendix.

\end{document}